\documentclass[a4paper]{article}
\usepackage{color,soul}

\usepackage{INTERSPEECH_v2,amssymb,amsmath,epsfig}
\usepackage[hidelinks]{hyperref}
\usepackage{microtype}
\usepackage{subfig}
\usepackage{tablefootnote}
\usepackage{xspace}

\title{Sequence-to-Sequence Models Can Directly Translate Foreign Speech}
\name{
  Ron J. Weiss$^1$,
  Jan Chorowski$^1$,
  Navdeep Jaitly$^{2\ast}$\thanks{$^\ast$ Work done at Google.},
  Yonghui Wu$^1$,
  Zhifeng Chen$^1$}
\address{
  $^1$Google Brain\\
  $^2$Nvidia}
\email{\{ronw,chorowski\}@google.com, njaitly@nvidia.com, \{yonghui,zhifengc\}@google.com}

\newcommand{\eg}{e.g.\xspace}
\newcommand{\etal}{et al.\xspace}

\newcommand{\fisherdev}{Fisher/dev\xspace}
\newcommand{\fishertest}{Fisher/test\xspace}

\RequirePackage[normalem]{ulem} %
\RequirePackage{color}\definecolor{RED}{rgb}{1,0,0}\definecolor{BLUE}{rgb}{0,0,1} %

\begin{document}

\maketitle
\begin{abstract}
We present a recurrent encoder-decoder deep neural network architecture that
directly translates speech in one language into text in another.  The model does
not explicitly transcribe the speech into text in the source language, nor does it
require supervision from the ground truth source language transcription during
training.  We apply a slightly modified sequence-to-sequence with attention
architecture that has previously been used for %
speech recognition
and show that it can be repurposed for this more complex task,
illustrating the power of attention-based models.

A single model trained end-to-end obtains state-of-the-art performance
on the Fisher Callhome Spanish-English speech translation task, outperforming a
cascade of independently trained sequence-to-sequence speech
recognition and machine translation models by 1.8 BLEU points on the Fisher test
set.  In addition, we find that making use of the training data in both
languages by multi-task training sequence-to-sequence speech translation and
recognition models with a shared encoder network can improve performance by a
further 1.4 BLEU points.
\end{abstract}

\section{Introduction} \label{sec:intro}
Sequence-to-sequence models were recently introduced as a powerful new method
for translation \cite{bahdanau2014neural, sutskever2014sequence}. Subsequently,
the model has been adapted and applied to various tasks such as image
captioning \cite{vinyals2015show,xu2015show}, pose prediction \cite{gkioxari2016chained},
and syntactic parsing \cite{vinyals2015grammar}.  It has also led to a new state of the
art in Neural Machine Translation (NMT) \cite{wu2016google}.
The model has also recently achieved promising results on automatic speech recognition (ASR)
even without the use of language models \cite{chorowski2015attention,chan2016listen,
zhang2016very,chorowski16towards}. These successes have only been possible
because the sequence-to-sequence (seq2seq) models can accurately model very complicated
probability distributions. This makes it possible to apply this model even
in situations where a precise analytical model is difficult to intuit.

In this paper we show that a single sequence-to-sequence model is powerful
enough to translate audio in one language directly into text in another language.
Using a model similar to Listen Attend and Spell (LAS) \cite{chan2016listen,zhang2016very}
we process log mel filterbank input features using a recurrent encoder. The encoder features
are then used, along with an attention model, to build a conditional next-step
prediction model for text in the target
domain. Unlike LAS, however, we
use the text in the translated domain as the target -- the source
language text is not used.

Conventionally, this task is performed by pipelining results from an ASR system
trained on the source language with a machine translation (MT) system trained to
translate text from the source language to text in the target language. However,
we motivate an end-to-end approach from several different angles.

First, by virtue of being an \emph{end-to-end} model, all the parameters are
jointly adjusted to optimize the results on the final goal. Training separate
speech recognition and translation models may lead to a situation where models
perform well individually, but do not
work well together because their error surfaces do not compose well. For example
typical errors in the ASR system may be such that they exacerbate
the errors of the translation model which has not been trained to see such errors
in the \emph{input} during training.  Another advantage of an end-to-end model is
that, during inference, a single model can have lower latency compared to a
cascade of two independent models. Additionally, end-to-end models have
advantages in low resource settings since they
can directly make use of
corpora where the audio is in one language while the transcript is in another.
This can arise, for example, in videos that have been captioned in other languages. It
can also reduce labeling budgets since speech would only need to be transcribed
in one language. In extreme cases where the source language does not have a writing
system, applying a separate ASR system would require first standardizing
the writing system -- a very significant
undertaking~\cite{bird2014collecting,anastasopoulos2016unsupervised}.

We make several interesting observations from experiments on
conversational Spanish to English speech translation.
  As with LAS models, we find that the model performs surprisingly well without
using independent language models in either the source or target language.
While the model performs well without seeing source language transcripts
during training, we find that we can
leverage them in a multi-task setting to improve performance.
Finally we show that the end-to-end model outperforms a cascade of
independent %
seq2seq ASR and NMT models.

\section{Related work}
Early work on \emph{speech translation} (ST) \cite{casacuberta2008recent} --
translating audio in one language into text in another -- used lattices from an
ASR system as inputs to translation models
\cite{ney1999speech,matusov2005integration}, giving the translation model access
to the speech recognition uncertainty.
Alternative approaches explicitly integrated acoustic and translation
models using a stochastic finite-state transducer which can decode the translated
text directly using Viterbi search \cite{vidal1997finite,casacuberta2004some}.

In this paper we compare our integrated model to results obtained from cascaded models on
a Spanish to English speech translation task \cite{post2013improved,kumar2014some,kumar2015coarse}.
These approaches also use ASR lattices as MT inputs.
Post \etal \cite{post2013improved} used a GMM-HMM %
ASR system.
Kumar \etal \cite{kumar2014some} later showed that using a better ASR
model improved overall ST results.  Subsequently
\cite{kumar2015coarse} showed that modeling features at the boundary of the
ASR and the MT system can further improve performance.
We carry
this notion much further by defining an end-to-end model for the entire task.

Other recent work on speech translation does not use ASR. %
Instead \cite{bansal2017towards} used an
unsupervised model to cluster repeated audio patterns which
are used to train a bag of words translation model.
In \cite{duong2016attentional} seq2seq models were used to align speech with
translated text, but not to directly predict the translations.
Our work is most similar to \cite{berard2016listen} which uses a LAS-like
model for ST on data synthesized using a text-to-speech system.
In contrast, %
we train on a much larger corpus composed of real speech.

\section{Sequence-to-sequence model} \label{sec:model}

We utilize a sequence-to-sequence with attention architecture similar to that
described in \cite{bahdanau2014neural}. The model is composed of three jointly
trained neural networks: a recurrent \emph{encoder} which transforms a sequence of input
feature frames $x_{1..T}$ into a sequence of hidden activations, $h_{1..L}$, optionally at
a slower time scale:
\begin{equation}
  h_l = \text{enc}(x_{1..T})
\end{equation}
The full encoded input sequence $h_{1..L}$ is consumed by a \emph{decoder} network
which emits a sequence of output tokens, $y_{1..K}$, via next step prediction:
emitting one output token (\eg word or character) per step, conditioned on the
token emitted at the previous time step as well as the entire encoded input
sequence:
\begin{equation}
  y_k  = \text{dec}(y_{k-1}, h_{1..L})
\end{equation}
The $\text{dec}$ function is implemented as a stacked recurrent neural network
with $D$ layers, which can be expanded as follows:
\begin{align}
  o^1_k,\, s^1_k   &= d^1(y_{k-1},   s^1_{k-1}, c_{k-1}) \\
  o^n_{k},\, s^n_k &= d^n(o^{n-1}_k, s^n_{k-1}, c_k)
\end{align}
where $d^n$ is a long short-term memory (LSTM) cell \cite{hochreiter1997long},
which emits an output vector $o^n$ into the following layer, and updates its
internal state $s^n$ at each time step.

\begin{figure*}[t]
  \centering
  \vskip-3ex
  \subfloat[Spanish speech recognition decoder attention.]{
    \includegraphics[height=0.33\paperheight]
                    {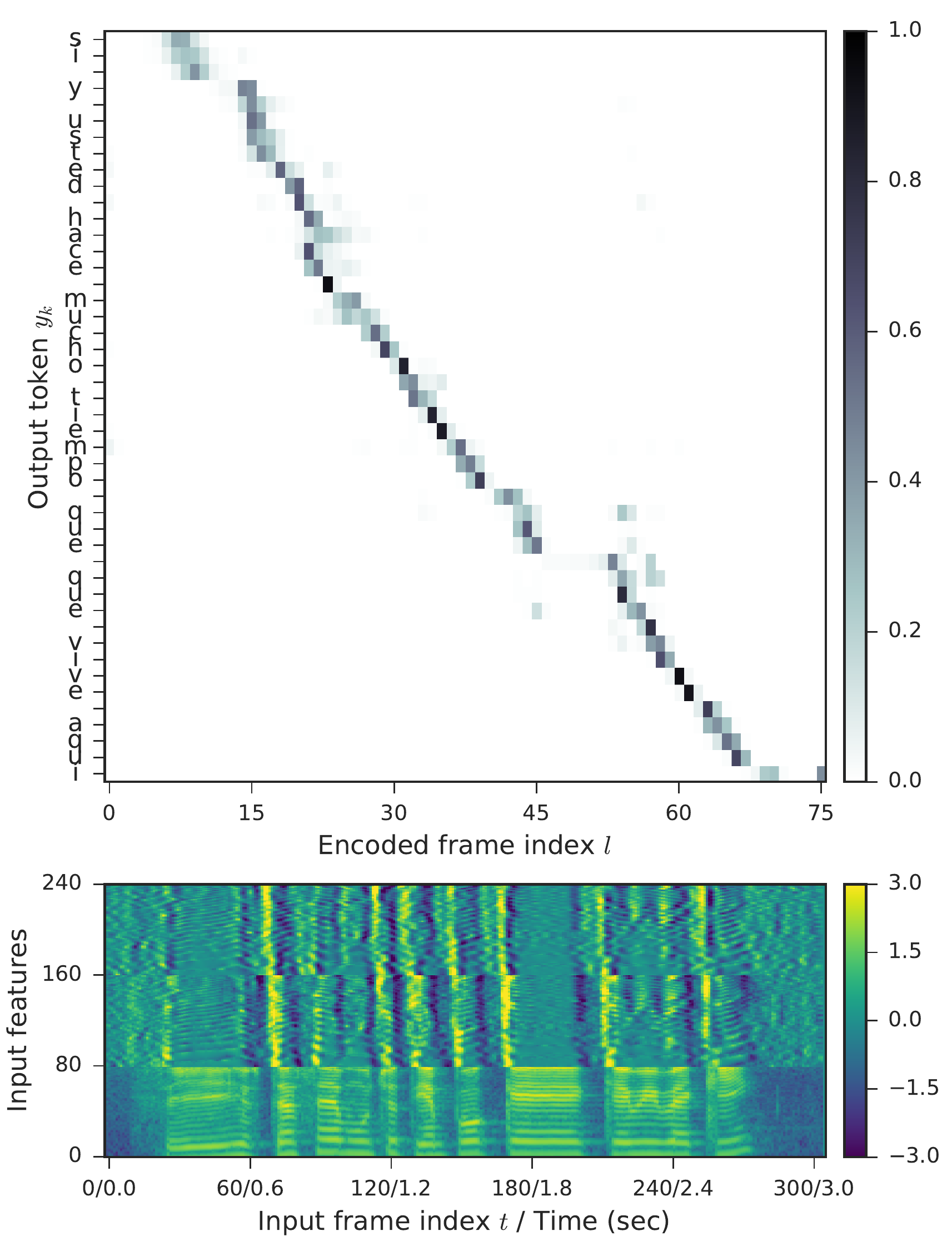}
    \label{fig:attention-asr}}
  \hfill
  \subfloat[Spanish-to-English speech translation decoder attention.]{
    \includegraphics[height=0.33\paperheight]
                    {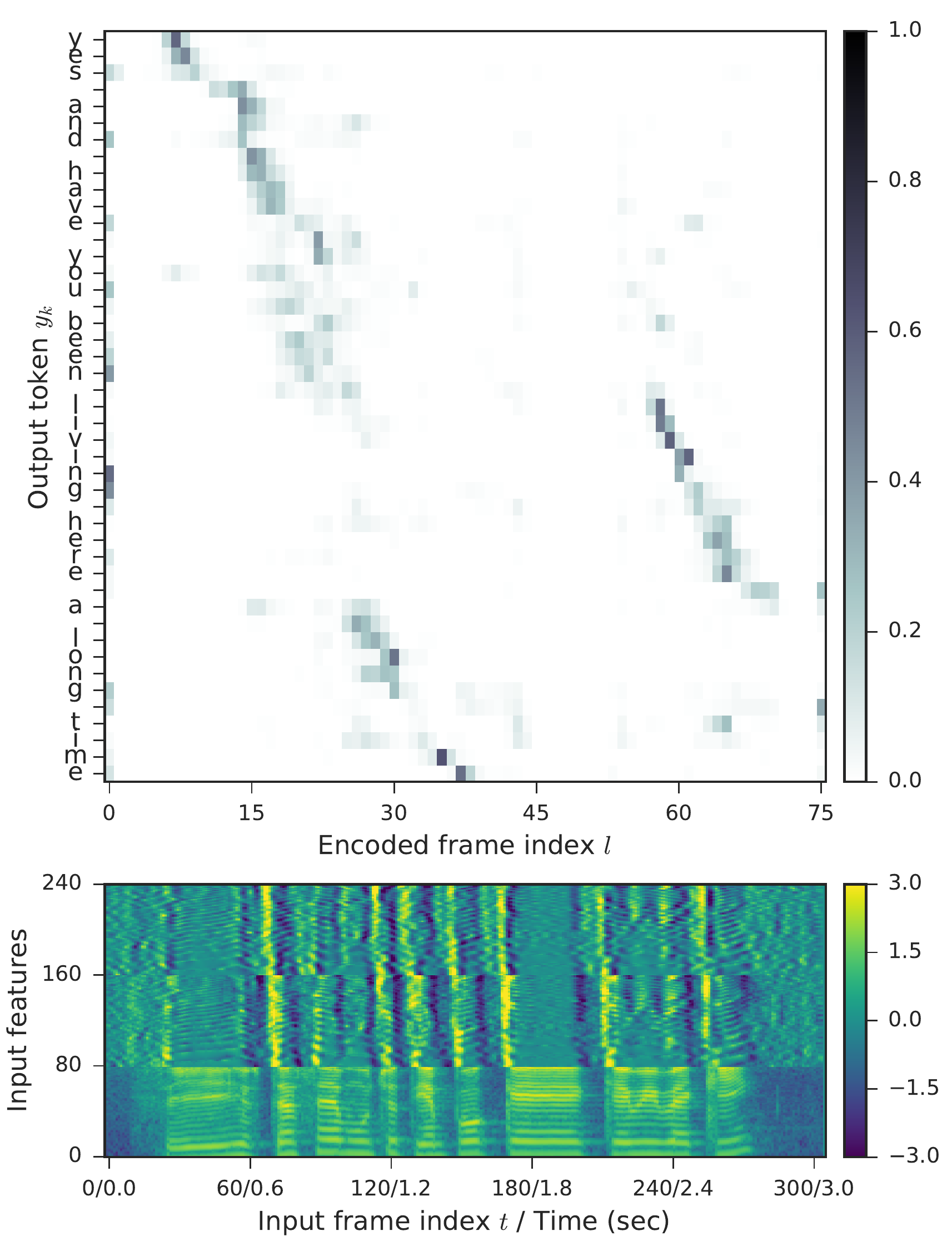}
    \label{fig:attention-ast}}
  \caption{Example attention probabilities $\alpha_{kl}$ from a multi-task model
    with two decoders.
    The ASR attention is roughly monotonic, whereas the translation attention
    contains an example of word reordering typical of seq2seq MT models,
    attending primarily to frames $l=58-70$ while emitting ``living here''.  The
    recognition decoder attends to these frames while emitting the corresponding
    Spanish phrase ``vive aqui''.
    The ASR decoder is also more confident than the translation attention,
    which tends to be smoothed out across many input frames for each
    output token. %
    This is a consequence of the ambiguous mapping between Spanish sounds and
    English translation.
  }
  \label{fig:attention}
\end{figure*}

The decoder's dependence on the input is mediated through an \emph{attention}
network which summarizes the entire input sequence as a fixed
dimensional context vector $c_k$ which is passed to all subsequent layers using
skip connections.
$c_k$ is computed from the first decoder layer output at each output
step $k$:
\begin{align}
  c_k &= {\textstyle\sum_l}\; \alpha_{kl} h_l \\
  \alpha_{kl} &= \text{softmax}(a_e(h_l)^T a_d(o^1_k))
  \label{eq:attention}
\end{align}
where $a_e$ and $a_d$ are small fully connected networks.  The $\alpha_{kl}$
probabilities compute a soft alignment between the input and output
sequences. An example is shown in Figure~\ref{fig:attention}.

Finally, an output symbol is sampled from a multinomial distribution computed
from the final decoder layer output: %
\begin{align}
  y_k &\sim \text{softmax}(W_y [o^D_k, c_k] + b_y)
\end{align}

\subsection{Speech model} \label{sec:speech_model}

We train seq2seq models for both end-to-end speech translation, and a baseline
model for speech recognition.  We found that the same architecture, a variation
of that from \cite{zhang2016very}, works well for both tasks.
We use 80 channel log mel filterbank features extracted from 25ms windows with a
hop size of 10ms, stacked with delta and delta-delta features.  The output
softmax of all models predicts one of 90 symbols, described in detail in
Section~\ref{sec:experiments}, that includes English and Spanish lowercase
letters.

The encoder is composed of a total of 8 layers.
The input features are organized as a $T \times 80 \times 3$ tensor, i.e. raw
features, deltas, and delta-deltas are concatenated along the 'depth' dimension.
This is passed into a stack of two convolutional layers with ReLU activations,
each consisting of 32 kernels with shape $3 \times 3 \times \text{depth}$ in
time $\times$ frequency.  These are both strided by $2 \times 2$, downsampling
the sequence in time by a total factor of 4, decreasing the computation
performed in the following layers.  Batch normalization \cite{ioffe2015batch} is applied
after each layer.

This downsampled feature sequence is then passed into a single
bidirectional convolutional LSTM
\cite{xingjian2015convolutional,bogun2015convlstm,zhang2016very} layer
using a $1 \times 3$ filter (i.e. convolving only across the frequency
dimension within each time step).  Finally, this is passed into a
stack of three bidirectional LSTM layers of size 256 in each
direction, interleaved with a 512-dimensional linear projection,
followed by batch normalization and a ReLU activation, to compute the
final 512-dimensional encoder representation, $h_l$.

The decoder input is created by concatenating a 64-dimensional embedding
for $y_{k-1}$, the symbol emitted at the previous time step, and the
512-dimensional attention context vector $c_k$.
The networks $a_e$ and $a_d$ used to compute $c_k$ (see
equation~\ref{eq:attention}) each contain a single hidden layer with 128 units.
This is passed into a stack of four unidirectional LSTM layers with 256 units.
Finally the concatenation of the attention context and LSTM output is passed
into a softmax layer which predicts the probability of emitting each symbol in
the output vocabulary.

The network contains 9.8m parameters.
We implement it with TensorFlow \cite{abadi2016tensorflow} and
train using teacher forcing on minibatches of 64
utterances.  We use asynchronous stochastic gradient descent across 10 replicas using the
Adam optimizer \cite{kingma2014adam} with $\beta_1= 0.9$, $\beta_2 = 0.999$, and
$\epsilon = 10^{-6}$.  The initial learning rate is set to $10^{-3}$ and decayed
by a factor of 10 after 1m steps.  L2 weight decay is used with a weight of
$10^{-6}$, and beginning from step 20k, Gaussian weight noise with std of 0.125 is
added to all LSTM weights and decoder embeddings.  We tuned all
hyperparameters to maximize performance on the \fisherdev set.

We decode using beam search with rank pruning at 8 hypotheses and a beam width
of 3, using the scoring function proposed in \cite{wu2016google}.  We do not
utilize any language models.  For the baseline ASR model we found
that neither length normalization nor the coverage penalty from
\cite{wu2016google} were needed, however it was helpful to permit emitting the
end-of-sequence token only when its log-probability was %
3.0 greater
than the next most probable token.  For speech translation we found that using
length normalization of 0.6 improved performance by 0.6 BLEU points.

\subsection{Neural machine translation model}
We also train a baseline seq2seq text machine translation model following
\cite{wu2016google}.  To reduce overfitting on the small training corpus we
significantly reduce the model size compared to those in \cite{wu2016google}.

The encoder network consists of four encoder layers (5 LSTM layers in total). As
in the base architecture, the bottom layer is a bidirectional LSTM and the
remaining layers are all unidirectional. The decoder network consists of 4
stacked LSTM layers. All encoder and decoder LSTM layers contain 512 units. The
attention network uses a single hidden layer with 512 units. We use the same
character-level vocabulary for input and output as the speech model described
above emits.

As in \cite{wu2016google} we apply dropout
\cite{srivastava2014dropout} with probability 0.2 during training to
reduce overfitting. We train using SGD with a single replica.
Training converges after about 100k steps using minibatches of 128
sentence pairs.

\subsection{Multi-task training}

Supervision from source language transcripts can be incorporated into
the speech translation model by co-training an auxiliary model with
shared parameters, \eg an ASR model
using a common encoder. This is equivalent to a multi-task
configuration \cite{luong2015multi}. We use the models and training
protocols described above with these modifications: we use 16
workers that randomly select a model to optimize at each step, we
introduce weight noise after 30k steps, and decay the learning rate
after 1.5m overall steps.

\section{Experiments} \label{sec:experiments}

We conduct experiments on the Spanish Fisher and Callhome
corpora of telephone conversations augmented with
English translations %
from \cite{post2013improved}.  We
split training utterances according to the provided segment annotations and
preprocess the Spanish transcriptions and English translations by lowercasing
and removing punctuation.
All models use a common set of 90 tokens to represent Spanish
and English characters, containing lowercase letters from both
alphabets, digits 0-9, space, punctuation marks,\footnote{Most are not used
  in practice since we remove punctuation aside from apostrophes from the
  training targets.} as well as special start-of-, end-of-sequence, and
unknown tokens. %

Following \cite{post2013improved,kumar2014some,kumar2015coarse}, we train all
models on the 163 hour Fisher train set, and tune hyperparameters %
on \fisherdev. We report speech recognition results as word
error rates (WER) and translation results using BLEU \cite{papineni2002bleu}
scores computed with the Moses toolkit\footnote{\url{http://www.statmt.org/moses/}}
\texttt{multi-bleu.pl} script, both on lowercase reference text after removing
punctuation.  We use the 4 provided Fisher reference translations
and a single reference for
Callhome.

\subsection{Tuning decoder depth}%

\begin{table}[t]
  \caption{Varying number of decoder layers in the speech translation
    model. BLEU score on the \fisherdev set.}
  \label{tbl:vary_ast_decoder}
  \centering
  \begin{tabular}{ccccc}
    \toprule
    \multicolumn{5}{c}{Num decoder layers $D$} \\
    1    &    2 &    3 &    4 &    5  \\
    \midrule
    43.8 & 45.1 & 45.2 & 45.5 & 45.3 \\
    \bottomrule
  \end{tabular}
\end{table}

It is common for ASR seq2seq models to use a shallow decoder, generally
comprised of one recurrent layer
\cite{chorowski2015attention,chan2016listen,zhang2016very}. In contrast,
seq2seq NMT models often use much deeper decoders,
\eg 8 layers in \cite{wu2016google}.
In analogy to a traditional ASR system, one may think of the seq2seq encoder
behaving as the acoustic model while the decoder acts as the language model.
The additional complexity of the translation task when compared to monolingual
language modeling motivates the use of a higher capacity decoder network.
We therefore experiment with varying the depth of the stack of LSTM layers used
in the decoder for speech translation and find that performance
improves as the decoder depth increases up to four layers, see
Table~\ref{tbl:vary_ast_decoder}.

Despite this intuition, we obtained similar improvements in performance on the
ASR task when increasing decoder depth, suggesting that tuning the
decoder architecture is worth further investigation in other speech settings.

\subsection{Tuning the multi-task model}
We compare two multi-task training strategies: one-to-many in which an
\emph{encoder} is shared between speech translation and recognition
tasks, and many-to-one in which a \emph{decoder} is shared between speech and
text translation tasks. We found the first strategy to perform better. %
We also found that performing updates more often on the speech translation task
yields the best results.  Specifically, we perform $75\%$ of training steps on
the core speech translation task, and the remainder on the auxiliary ASR
task.

\begin{table}[t]
  \caption{Varying the number of shared encoder LSTM layers in the
    multi-task setting. BLEU score on the \fisherdev set.}
  \label{tbl:mt_sharing}
  \centering
  \begin{tabular}{ccccc}
    \toprule
    \multicolumn{4}{c}{Num shared encoder LSTM layers} \\
    3 (all) &    2 &    1 &    0 \\
    \midrule
    46.2    & 45.1 & 45.3 & 44.2 \\
    \bottomrule
  \end{tabular}
\end{table}

Finally, we vary how much of the encoder network parameters are shared across
tasks.  Intuitively we expect that layers near the input will be less sensitive
to the final classification task, so we always share all encoder layers through
the conv LSTM but vary the amount of sharing in the final stack of LSTM layers.
As shown in Table~\ref{tbl:mt_sharing} we found that sharing all layers
of the encoder yields the best performance.  This suggests that the encoder
learns to transform speech into a consistent interlingual subword unit
representation, which the respective decoders are able to assemble into phrases
in either language.

\subsection{Baseline models}%

\begin{table}[t]
  \caption{Speech recognition model performance in WER.}
  \label{tbl:asr_results}
  \centering
  \begin{tabular}{lccc@{\hskip1em}c@{\hskip1ex}c}
    \toprule
    & \multicolumn{3}{c}{Fisher} & \multicolumn{2}{c}{Callhome} \\
         &  dev & dev2 & test         & devtest & evltest \\
    \midrule
    Ours\tablefootnote{\label{note:average}Averaged over three runs.}
         & 25.7 & 25.1 & 23.2         & 44.5 & 45.3 \\
    Post \etal \cite{post2013improved}
         & 41.3 & 40.0 & 36.5         & 64.7 & 65.3 \\
    Kumar \etal \cite{kumar2015coarse}
         & 29.8 & 29.8 & 25.3         &  --  & -- \\
    \bottomrule
  \end{tabular}
\end{table}

\begin{table}[t]
  \caption{Translation BLEU score on ground truth %
    transcripts.}
  \label{tbl:mt_results}
  \centering
  \begin{tabular}{lccc@{\hskip1em}c@{\hskip1ex}c}
    \toprule
    & \multicolumn{3}{c}{Fisher} & \multicolumn{2}{c}{Callhome} \\
    &  dev & dev2 & test         & devtest & evltest \\
    \midrule
    Ours & 58.7 & 59.9 & 57.9    & 28.2 & 27.9 \\
    Post \etal \cite{post2013improved}
         &  --  &  --  & 58.7    &  --  & 27.8 \\
    Kumar \etal \cite{kumar2015coarse} %
         &  --  & 65.4 & 62.9    &  --  & -- \\
    \bottomrule
  \end{tabular}
\end{table}

We construct a baseline %
cascade of a Spanish ASR seq2seq model whose output is passed into a
Spanish to English NMT model.
Our seq2seq ASR model attains state-of-the-art performance on the Fisher and
Callhome datasets compared to previously reported results with HMM-GMM
\cite{post2013improved} and DNN-HMM \cite{kumar2015coarse} systems, as shown in
Table~\ref{tbl:asr_results}.
Performance on the Fisher task is significantly better than on Callhome since it
contains more formal speech, consisting of conversations between strangers
while Callhome conversations were often between family members.

In contrast, our MT model slightly underperforms compared to previously reported
results using phrase-based translation systems
\cite{post2013improved,kumar2014some,kumar2015coarse} as shown in
Table~\ref{tbl:mt_results}.  This may be because the amount of training data in the
Fisher corpus is much smaller than is typically used for training NMT systems.
Additionally, our models used characters as training targets instead of word-
and phrase-level tokens often used in machine translation systems, making them
more vulnerable to \eg spelling errors.

\subsection{Speech translation}

\begin{table}[t]
  \caption{Speech translation model performance in BLEU score.}
  \label{tbl:st_results}
  \centering
  \begin{tabular}{l@{\hskip1ex}ccc@{\hskip1.5ex}c@{\hskip1ex}c}
    \toprule
          & \multicolumn{3}{c}{Fisher} & \multicolumn{2}{c}{Callhome} \\
    Model &  dev & dev2 & test         & devtest & evltest \\
    \midrule
    End-to-end ST\textsuperscript{\ref{note:average}}
          & 46.5 &  47.3  & 47.3       &    16.4 &    16.6 \\
    Multi-task ST / ASR\textsuperscript{\ref{note:average}}
          & 48.3 &  49.1  & 48.7       &    16.8 &    17.4 \\
    \midrule
    ASR$\rightarrow$NMT cascade\textsuperscript{\ref{note:average}}
          & 45.1 &  46.1  & 45.5       &    16.2 &    16.6 \\
    Post \etal \cite{post2013improved} & -- & 35.4 & --   & -- & 11.7 \\
    Kumar \etal \cite{kumar2015coarse} & -- & 40.1 & 40.4 & -- & -- \\
    \bottomrule
  \end{tabular}
\end{table}

Table~\ref{tbl:st_results} compares performance of different systems on the full
speech translation task.  Despite not having access to source language
transcripts at any stage of the training, the end-to-end model
outperforms the baseline cascade, which passes the 1-best Spanish ASR output into
the NMT model, by about 1.8 BLEU points on the \fishertest set.
We obtain an additional improvement of 1.4 BLEU points or more on all Fisher datasets in the
multi-task configuration, in which the Spanish transcripts are used for
additional supervision by sharing a single encoder sub-network across independent
ASR and ST decoders.
The ASR model converged after four days of training (1.5m steps), while the ST
and multitask models continued to improve, with the final 1.2 BLEU point
improvement taking two more weeks.

Informal inspection of cascade system outputs yields many examples
of compounding errors, where the ASR model makes an insertion or deletion that
significantly alters the meaning of the sentence and the NMT model has no way
to recover.  This illustrates a key advantage of the end-to-end approach
where the translation decoder can access the full latent representation of
the speech without first collapsing to an
n-best list of hypotheses.

A large performance gap of about 10 BLEU points remains between these results and
those from Table~\ref{tbl:mt_results} which assume perfect ASR, indicating
significant room for improvement in the %
acoustic modeling component of the speech translation task.

\section{Conclusion} \label{sec:conclusion}

We present a model that directly translates speech into text in a
different language.  One of its striking characteristics is that its
architecture is essentially the same as that of an attention-based ASR
neural system. Direct speech-to-text translation happens in the same
computational footprint as speech recognition -- the ASR and
end-to-end ST models have the same number of parameters, and utilize
the same decoding algorithm -- narrow beam search. The end-to-end
trained model outperforms an ASR-MT cascade even though it never
explicitly searches over transcriptions in the source language during
decoding.

While we can interpret the proposed model's encoder and decoder
networks, respectively, as acoustic and translation models, it does
not have an explicit concept of source transcription. The two
sub-networks exchange information as abstract high-dimensional real
valued vectors rather than discrete transcription lattices as in
traditional systems.  In fact, reading out transcriptions in the
source language from this abstract representation requires a separate
decoder network.  We find that jointly training decoder networks for
multiple languages regularizes the encoder and improves overall speech
translation performance.
An interesting extension would be to construct a multilingual speech
translation system following \cite{johnson2016google} in which a
single decoder is shared across multiple languages, passing a
discrete input token into the network to select the desired output
language.

\bibliographystyle{IEEEtran}
\bibliography{refs}

\end{document}